# Touch in Human Social Robot Interaction: Systematic Literature Review with PRISMA method

Christiana Tsirka[1], Anna-Maria Velentza '[0000-0002-1251-571X]'[,1,2,], Nikolaos Fachantidis '[0000-0002-8838-8091]' [1, 2]

[1]School of School of Educational and Social Sciences, University of Macedonia,GR

[2] Laboratory of Informatics & Robotic Applications in Education & Society (LIRES), University of Macedonia, GR

Corresponding Author: Anna-Maria Velentza, annamarakiv@gmail.com

Contributing Authors: christianatsirka2@gmail.com, nfachantidis@uom.edu.gr,

# Touch in Human Social Robot Interaction: Systematic Literature Review with PRISMA method

Abstract

In the past two decades, there has been a continuous rise in the deployment of robots fulfilling social roles that expands across various industries such as guides, service providers, and educators. To establish robots as integral allies in daily life, it is essential for them to deliver positive and trustworthy experiences, achieved through seamless and satisfying interactions across diverse modalities and communication channels. In the realm of human-robot interactions, touch plays a pivotal role in facilitating meaningful connections and communication. To delve into the significance of haptic technologies and their impact on interactions between humans and social robots, an exploration of the existing literature is essential, since the research about touch is the most underrepresented between the other communication channels (facial expressions, movements, vocals etc). A systematic literature review has been carried out, identifying 42 articles with the Preferred Reporting Items for Systematic Reviews and Meta-Analyses (PRISMA), related to touch and haptic technologies and interaction between humans and social robots in the twenty years (2001 -2023). The results show the main differences, pros and cons between the materials and technologies that has been primary used so far, the qualitative and quantitative research that links the HRI touch studies with the human emotion and also the types of touch and repeatability of those methods. The study identifies research gaps and outlines future directions, while serves as a guide for anyone who will be interesting in conducting HRI touch research or build a haptic system for a social robot.

## 1. Introduction

The field of social robotics has experienced significant growth, with diverse computer science laboratories across companies and universities dedicated to optimizing the form, functions, and communication frameworks of social robots for achieving optimal Human-Robot Interaction (HRI)[1]. These social robots are designed to engage with humans in a socio-emotional manner, and some operate autonomously, enabling interactions with individuals, other robots, and their environment [2].

The development of social robots revolves around the concept of designing machines capable of natural and seamless communication and collaboration with humans [3]. Some social robots possess the capacity to spontaneously initiate interactions with people [3]. These social robots extend beyond task execution, engaging with people in ways that cater to individual preferences, needs, personal experiences, and cultural backgrounds [4]. While humanoid robots have gained attention, non-humanoid designs, such as animal or cartoon-like robots, may be better received by children and adults alike [5].

As technological advances have allowed for more detailed mimicry of human social characteristics, the question arises regarding the extent to which human attributes can be instilled in social robots[6]. The widespread integration of social robots into daily life and increased acceptance among people is expected [5]. Among the most interesting aspects of human-robot contact is haptic communication, in which touch is a key way of showing intimacy[7].

Human-robot haptic interaction is a multidisciplinary field, involving Psychology, Neurophysiology, Engineering, and Computer Science, and demands suitable tactile interfaces, entailing precise material selection and sensor technologies[8],[9],[10]. Since social robots will soon be used in many aspects of daily life, it is very important to study how people interact with them physically. The role of touch and affection in HRI has been extensively discussed [11] for care settings, during the ability to show affection [7]. On some late papers, the researchers focus is to leverage latest technologies to design low cost touch systems [12], while on others on how to design more inclusive appearance robots by using fur and soft materials[13].

The purpose of this work is to systematically bring together data from related studies in order to fully classify interactions between humans and social robots, which will shed light on this complicated area.

## 2. Research Methodology

For this review we followed the Preferred Reporting Items for Systematic Reviews and Meta-Analyses (PRISMA) guidelines[14]. Google Scholar search engine was initially used to conduct a comprehensive search and identify pertinent scholarly papers. The selection criteria for articles encompass the timeline from 2001 and date this research was conducted, 2023. Although the field is relatively new, we decided to search for older articles and technologies that may lead to a fruitful conversation regarding the efficiency or methodology of articles first published in the field. During the initial step, the process of locating relevant articles was conducted using targeted keywords. The keywords used were about haptic technologies, namely 'haptic' OR 'tactile', OR 'touch' AND 'social robot*'

The inclusion criteria for this study were selecting articles based on their title and/or abstract. Specifically, the focus was on studies that examined technologies and materials utilized in human-social robot physical contact, different types of touch, emotions, and factors connected to interaction timing, i.e., duration of touch. Moreover, the articles should be peer-reviewed journals or conferences also classified in one of the following citation or research database: Scopus, Web of Science, ACM Digital Library, IEEE Xplore Digital Library, and Science Direct.

As previously stated, the primary subject of this work is the social interaction between robots and humans. Consequently, we excluded papers dedicated to robotic systems or interactions with individuals lacking social features or not assuming social roles during those interactions. A total of 139 articles were identified and subsequently chosen for further analysis, 62 articles by conducting a search using the keywords ‘haptic’ AND ‘social robot*’, 34 articles by ‘tactile’ AND ‘social robot*’, and 43 by ‘touch’ AND ‘social robot*’.

The chosen papers were systematically archived by assigning them a relevant tag. Subsequently, the articles associated with each tag were exported to a .csv file. Each comma-separated values (CSV) file contains data pertaining to authors, title, publication details, page count, publication date, and publisher name for individual articles. After excluding duplicates, we conducted a quality assessment. By employing this methodology, we could discern the articles' applicability to the subject under review and their relevance to the evaluation of robot haptic technology in the context of human-social robot interaction. For this purpose, were formulated new tags based on the content of the article. Due to the inherent characteristics of this form of experimentation, the significance of sample size or replicability in determining quality has not been established. Because of the potential existence of diverse application protocols, the sample sizes utilized in the various experiments vary considerably.

In the subsequent stage, five novel theme tags were generated considering the content of the gathered articles. The designated categories included technology, materials, type of touch, sensations, duration, and repetition. Upon conducting a search pertaining to the five tags, it was discovered that 35 studies were associated with technology, 32 studies were focused on materials, 18 studies explored the concept of touch, 18 studies delved into emotions, and 9 research examined the aspect of duration/repetition. The items were saved within their corresponding tags. Subsequently, the articles corresponding to each tag were extracted and saved in a .csv file.

In the conducted trials wherein the 5 novel tags were implemented, an additional tag referred to as "phase 2" was introduced. By exporting the .csv file with the tag labeled "phase 2," it became feasible to ascertain the commonalities among articles marked with technology, materials, type of contact, sentiments, duration, and repetition. There was a total of 42 articles contained within the .csv file designated as phase 2. After deducting a total of 42 articles from the initial count of 109 articles during the initial phase of the second stage, it was determined that there were 67 articles that were common to both phases. A total of 67 papers were excluded from the analysis when it was determined that they were assigned in more than one tags. Hence, the review incorporated a total of 42 studies. The creation of the PRISMA flowchart depicted in Fig.1 provides a schematic representation of the analysis, leading up to the final count of studies included in the review.

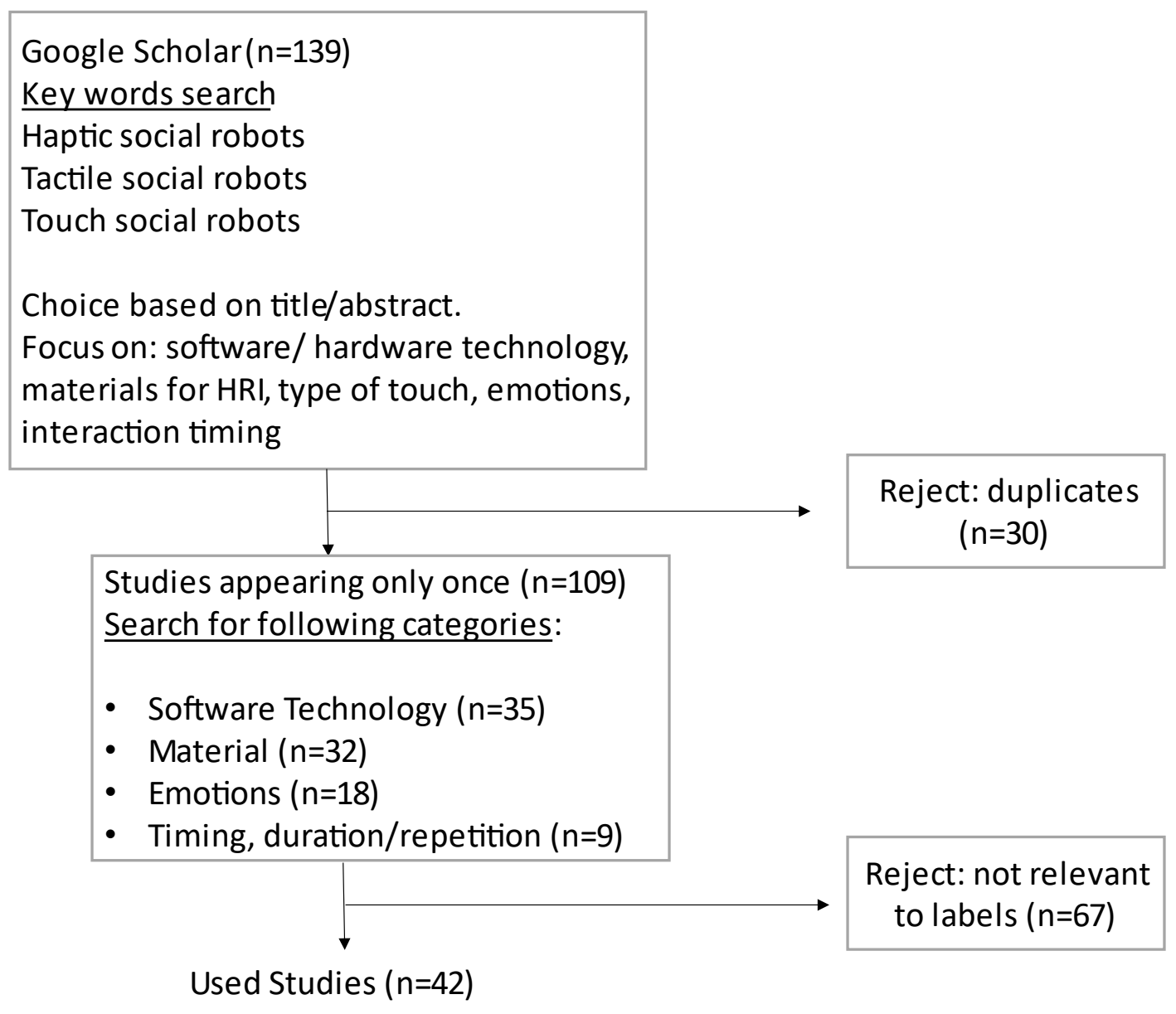

**Fig.1**. Prisma Flowchart

## 3. Analysing Categories

### *3.1 Technology and touch surface*

In the field of technology and touch surface the following common features were identified. From this categorization we can see that the soft materials are being preferred, with robots covered by fur or fabric. The appearance and the elements of cuteness also play an important role with a lot of robots imitating animal figures and features. The humanoid robots Nao and Pepper are also being used. Technologies such as conductive material sensors and accelerometers have been leveraged to enhance the interaction.

#### *3.1.1 Texture*

3.1.1.1 Fur and conductive fur [15], [16], [17], [18], [19], [20], [21], and [22]

The Haptic Creature is a touch-centric social robot, and furry animal-like robot. The robot is clothed with faux fur on a fiberglass[18] and humans interacting with it were instructed to touch the fur in different ways such as by gripping or pressuring it [15], exploring the basic principles of emotional touch[20]. Flagg et al., [17] present a new type of sensor based on conductive fur, which is sensitive to movements not available in conventional pressure sensors. The sensor registers movement by measuring the changing current from the conductive threads of the fur. They also conductive fur, Styrofoam interior and plastic frame for another of their studies [22]. Stiehl, et al., [21] study how a teddy bear with touch sensors can be used to treat patients. Moreover, Hefter, et al., [19], designed a system that combines data from a fur fabric that detects when and how it is being handled, with information from a vision-based emotion recognition system. The fur sensor's voltage is measured by an Arduino, which then converts it to the proper value and serial communication is used to send the values to the computer as they are read.

3.1.1.2 Fabric [23], [24], [25], [26], [27]

Urethane gel, urethane sponge, and a thin fabric glove are the materials utilized in the palm of the arm [26], while in [24] the act of handshake is accompanied by the presence of a glove composed by fabric elastic material. The robotic system that enables close communication over long distances is the subject of [25]'s study. The physical form exhibits curvilinear contours that imply the presence of a cranial region and a pair of lower extremities. The item is adorned with plush and silk textiles that exude a sense of intimacy and are well-suited for domestic settings. HuggieBot, in order to simulate the sense of touch, the robot's midsection incorporates two airbags, while the user interacts with tactually responsive surfaces that are appropriately clothed [23]. Additionally, the system described in [27] was developed to identify tactile objects using an EeonTex conductive fabric sensor and an Arduino Mega microprocessor. This was achieved by utilizing the analog input pins and digital output pins of the CPU.

*3.1.1.3 Surface type for skin that changes texture*: [28], [29], [30]

A mold-casting method is used to make the main body of the texture skin out of an elastic material that can be stretched. 3D-printed PLA is used to make a plastic positive of the inner cavity and fluidic channels, which is then set in a mold box. The bottom layer is made with a film that can't be stretched inside it, and the cured top layer is connected to the bottom layer to make the cavities that are closed. One part is used to form the outside shape of the TU-GB/TU-S mixed module, and the other part is used to make the fluidic cavity and chamber. We insert the rigid parts of the spikes before casting the top layer[28], [29]. The robot shell is covered with actuated small smooth or pointed bumps, with programmable frequency and amplitude patterns. In a controlled study (n = 139) [28]. In their research, [30] investigate the functionalities of interactive robotic skin by drawing inspiration from biological organisms. They approach this subject from four distinct angles: expression, perception, regulation, and mechanical action. The researchers showcase six prototypes of skin with the ability to change texture and delve into their potential for expression [30].

*3.1.2 Appearance*

*3.1.2.1 Robots that look like fur toy animals* [15], [31], [16], [18], [32], [27], [21], [25], [33], [34], [22]

An inflatable vest called "Hug Over a Distance" can be remotely triggered to produce a hug-like feeling. In other words, one person utilizes the vest while the other employs a koala bear that he touches on the tummy. The vest inflates upon contact with the koala, creating the impression of a hug [33]. The robot 'The Hug', also replicates a toy-like appearance [25], and [34], "Huggy Pajama", also designed for distance hug serving parent- child communication looks like a fur teddy bear. Haptic Creature was also based on animal like characteristics [15], [16], [18], while [22], and [15] made the Nao robot look like a fur animal, by dressing it up look like a koala. Shiomi et al., [32] implemented a robot teddy bear in human size and [27] a robot that looks like a small animal. Similarly, [21] created an animal- like effect by adding a under a soft cover of silicone skin and fur fabric.

*3.1.2.2 Humanoid Robots*

The Robots *Nao* [31], [35], and [9] and *Pepper* has [8] and [36] has been used for research purposes. In some cases, it is hard to specify the limits between humanoid and animal like appearance, like the [31] research, in which the Nao robot has been deployed koala-dressesd. The robot uses fabric-based touch sensors to detect and distinguish various touches to aid children with autism (ASD) and their families [31]. Willemse and Van Erp [35] studied whether Nao robot touches could elicit good responses in human users like human touches to improve human-like social robot interactions and if a favorable social bond with the robot affects these responses. Andreasson, et al. [9] chose to research gender differences in Nao robot tactile interaction feelings, where participants touched a miniature humanoid robot to express eight emotions.

Hirano, et al. [8], and earlier, [36] examine human-social robot Pepper tactile interaction.

[37] introduce Maggie, a 1.35-meter humanoid robot, aiming to support home chores, medical care for the elderly, people with mobility or cognitive limitations, entertainment, education, personal assistance, delivering instructions at public information sites, and more, also by employing capacitive sensors for hard-shelled robots. The humanoid robot i-cub has also been deployed for touch studies[38].

*3.1.2.3 e-skin* [39], and [40]

The primary challenges examined in [40] stem from energy autonomy, the handling and transmission of extensive data sets, neural-inspired methodologies, and fabrication techniques for achieving high-performance soft electronic skin. These challenges also encompass advancements in the domain of flexible and printable electronics. Saadatian et al., [39], in order to create the sense of kiss with the ‘Kissinger’, tried to closely replicate the shape and color of human lips, utilizing RTV 560 specialized silicone material.

*3.1.3 Technology enhancing interaction.*

*3.1.3.1 Capacitative technology* [41], [37], [42], [43], [44], [45], and [12]

Lin et al., [41] develop a low-cost tactile interface system for social robots that allows full body involvement, employing capacitative technology. Salichs, et al. [37] also introduce a capacitative technology touch pattern for emotion recognition system. Koo et al., [45] uses capacitive touch and accelerometers for hard-shelled robots like [44] and [12]. However, they evaluate system performance. In contrast to [44], [12] evaluate the system's operational correctness, but not its use of data to determine user emotions. [42] created a capacitive mechanism to help a social robot turn its head with copper foil touch surfaces on a table that makes the system simple and easy to use. Silvera Tawil, et al., [43] decipher touch patterns on an artificial arm with unique skin with 19 round electrodes and soft fabric. This work introduces thin, flexible, elastic artificial skin that can detect touch position, duration, and intensity with the aid of capacitative technology [43].

*3.1.3.2 Sensors FSR* [15], [16], [46], [18], [20], and [39]

[30] present a haptic interaction system for social robots based on a 1.5” touch pad using Force Sensing Resistor (FSR) technology. In another study [44] are also using a FSR resistance network. The restricted coverage of FSRs limits detection possibilities. Furthermore, tangential forces are involved in many touches, although FSRs are sensitive to forces applied perpendicular to the surface [20]. Additionally, the researchers of ‘Kissinger’, in order to detect the kiss, employed FSR [39]. Moreover, Haptic creature operates with Sensors FSR [15], [16], [18].

*3.1.3.3 Accelerometer*: [12], [15], [16], [18], [24], [20], [44], [45]

In [44] the accelerometer is capable of detecting both the robot's motion and the slight vibrations brought on by a touch. An accelerometer is also used by the Haptic Creature [15], [16], and [18]. In their experiment, [24] examine the pressure individuals apply during a particular touch-based social interaction—the handshake—based on their emotional state. Eleven participants were given a set of handshakes to practice every day. Twenty-three piezo-resistive sensors were incorporated into custom gloves, and an accelerometer was also added [24]. Accelerometer has been also used in KaMERo [44], which is being developed for more sophisticated human interaction in entertainment applications. An accelerometer has also been placed on the hard-cover surface of both [45] and [12] to record the vibrations transmitted by the shell during interaction with the system.

*3.1.3.4 Temperature and heat*: [43], [47], [21], [25]

Silvera-Tawil, et., al. [43] explore haptic communication in relation to the human sense of touch, examining its nuances based on gender and specific body parts. This investigation contributes to understanding haptic interactions in Human-Robot Interaction (HRI), incorporating temperature sensors among other elements. In [21]'s robotic design, a comprehensive array of sensors, including a temperature sensor, is integrated beneath a soft cover composed of silicone skin and fur fabric. This combination enhances the robot's tactile capabilities. Additionally, [47] present a robot featuring variable softness achieved through a heat-responsive gel, showcasing the integration of thermal responsiveness in robotic design. [25] introduces "The Hug," where touch sensors on the back enable the transmission of movements between robots. These movements are then translated into unique vibration patterns in the arms and abdomen through integrated motors. The thermal fibers surrounding the belly contribute to a gradual increase in warmth, enhancing the overall tactile experience [25].

*3.1.3.5 Measurement of pressure values in N and in kPa*: [48], [49], [46], [24] and [34]

In the study of [48], leveraging a 3-axis Hall Effect sensor enables the detection of magnetic field variations across three axes, allows the sensor to ascertain both the magnitude and direction of the force in a three-dimensional space. These sensors exhibit the capability to identify forces at levels as minimal as 10 mN. [46] and [34] are also measuring the presure in N, on their exposed sensors [30] and their huggy pajama teddy bear [34] accordingly.  Measurement of sensor pressure values in kPa has been also demonstrated on the cine-skin silicon touch of [49], while [24] who focus on handshakes and thus the pressure is a crucial factor for the social interaction between the robot and the humans also choose the same measurement.

### *3.2 Type of touch*

*3.2.1 Studies with different types of touch* [41], [50], [49], [15], [51], [31], [17], [46], [8], [35], [18], [19], [38], [36], [27], [43], [21], [44], [45], [22]

The research conducted by [41] explores various facets of the haptic perception system, encompassing the development of an economical haptic sensor array for the entire body of a social robot. Simultaneously, [15], [18], examine interactions between humans and the Haptic Creature. Vasara and Surakka [46] investigate the impact of different facial expressions on diverse haptic behaviors, while [8] and [36] explore how communication cues influence impressions of human-robot touch interaction. Regarding the touch behavior, two approaches were chosen, one is touching the robot, and the other is being touched by the robot [8], [36]. Additionally, [35] investigate whether robot-initiated touches evoke responses comparable to human touch and whether pre-existing positive social bonds with the robot modulate these responses. [38] validates their method through experiments on tactile interaction in simulated and real environments.

Burns et al., [31] introduce a touch-aware robot for children with autism, customizing the robot's response system based on the child's touches through input from a therapist or caregiver. [27] opt for a low-cost touch sensor to collect touch data, revealing users' unique "touch signatures" that enable the robot to recognize individuals. [43] employ an algorithm to classify eight types of touch, achieving a 71% accuracy rate in trials, comparable to human touch recognition. Moreover, gender and cultural background were examined and found to have no statistically significant effect on classification results[43]. The study of  [49] presents a sensor capturing the entire frame interaction, utilizing a self-organizing neural network for touch recognition and measurement. Additionally, Huisman [51] categorizes touch types, touch points, and corresponding technologies, offering reflections on the current state and future directions of touch technology in social robots.

Furthermore, [50] introduce a learning system for touch type detection and recognition, utilizing three contact microphones within the robot for sensing touch-induced sound signals, which are then processed using Machine Learning techniques. This approach, demonstrated on social robots Maggie and Mini, allows the robot to sense touches, recognize touch types, and estimate touch locations, highlighting the advantage of using contact microphones to cover entire solid parts of the robot with minimal sensors [50].

The objective of [19] is to develop a socially responsive robot that conveys sympathy, fostering a positive user perception. This innovative system integrates two distinct input devices, touch and a camera, to assess users' emotional states and deliver suitable responses [19]. Furthermore, [17], [22] employ a sensor embedded in fur threads that utilizes machine learning for touch classification, achieving an impressive recognition accuracy of 82%, as validated by an evaluation involving seven participants. The recognition of touch behaviors around the identification of various touches, is also one of the major objectives of [21], [44], and [45].

#### *3.2.2 Physical Hug* ***[23]****,* ***[32]*** *and Hug from Distance* ***[25]****,* ***[33]*** and ***[34]***

Hugs are a social way of expressing intimacy and [32] and [23] highlight the significance of hugging as a mode of human interaction with a social robot, employing a life-sized teddy bear as a surrogate. The question of whether these physical interactions yield comparable results within the context of human-robot interaction remains relatively uncharted, despite the discovery of numerous positive effects associated with tactile engagements. Consequently, they employ a specially devised hugging robot and engage in empirical investigations to explore the physical interactions that contribute to fostering engagement. Moreover, [27] suggests six guidelines for creating authentic and enjoyable robotic hugs: the robotic hugger should possess a soft and warm texture, be of human size, visually perceive its wearer, adapt the hug's size and position to the wearer, and discern when the wearer wishes to conclude the hug.

In the realm of remote hugging research, [25] analyse the subject using the robot known as "The Hug." The researchers of this study stress that while advancements in robotics offer powerful technology, it is equally essential to explore aspects like shape, materials, and behavior. This exploration aims to strike a balance between human needs, technological capabilities, and the context of usage, particularly in supporting intimate communication across distances. Similarly, [33], following the concept of distant hugging, address the challenge of individuals in close relationships separated by physical distance who struggle to convey their emotions effectively. They introduce the "Hug Over a Distance" system, featuring an inflatable vest that can be activated remotely, creating a sensation akin to a hug. In this setup, one user wears the vest, while the other employs a koala bear, stroking its belly, triggering the inflation of the vest to simulate a hug. Teh, et al.,[34] present a concept that aligns with the principles of the "Hug Over a Distance" system, known as the "Huggy Pajama." This system revolves around remote hugging intended for parent-child communication and employs a plush teddy bear as the initiating device, much like the koala used in the "Hug Over a Distance" approach.

***3.2.3*** Kiss [39]

The act of kissing is intricately linked with emotional expressions, and Kissenger [39] is specifically crafted to enhance prevailing telecommunications technologies. The challenges in the design and development of this innovative device are met through an iterative design process, incorporating ongoing user evaluation at each stage. To ensure effectiveness, devices undergo comprehensive assessment in both short- and long-term user studies, particularly focusing on couples engaged in long-distance relationships.

***3.2.4*** *Handshake*: [48], [24], [26] and [52]

The investigation of [48] focus into the dynamics of human-social robot interaction through handshakes, employing touch sensors and the Vizzy robot serves as the robotic platform for this study, specifically designed to act as a human assistant in social interaction scenarios. Similar to the approach in [48], researchers in [24] also explore the realm of human-robot contact through handshakes, emphasizing the notable advantages of touch in interpersonal social interaction. This underscores the importance of equipping social robots with emotional tactile capabilities. However, most studies concentrate on the outcomes and subjective interpretation of touch rather than investigating the specific methods and techniques employed in the act of touch. Understanding tactile perception enables the emulation of natural behavior in robot manipulation and aids in identifying the user's intentions during physical interaction with the robot.

In a parallel investigation, another group of researchers tackles the handshake challenge using a setup featuring a screen displaying a remote user and an arm resembling a human hand [26]. They observe that studies on tactile and visual communication often follow separate paths, leaving uncertainties about the potential enhancement of communication through touch when video and audio channels are already available. To explore this, they analyze remote handshakes, where a robotic hand beneath the video conferencing terminal's screen mimics the opening and closing motion of an interlocutor's hand [26]. Additionally, [52] explore handshakes with a robot incorporating facial expressions and their experiment showcases how participants combine facial expressions and haptic feedback to perceive emotions during interactions with an expressive humanoid robot.

### *3.3 Emotions*

*3.3.1 Evaluation of emotions in touch behavior* [48], [28], [15], [31], [23], [29], [46], [35], [18], [19], [38], [37], [24], [30], [20], [9], [21], [44], [10], [22], and [52]

Participants in [9] communicated eight emotions to a small humanoid robot using touch, and distinctive variations were identified among the emotions. This allowed for the classification of emotions based on the intensity of the conveyed emotion, considering both touch and duration[9]. Introducing their system on the KaMERo robot, [44] designed for advanced interactions in entertainment applications, utilizing touch patterns to enable the robot to comprehend human rewards or punishments within the context of a game.

Robot-initiated touches in [35] can induce positive responses without extensive prior bonding. In the handshakes performed in [24], there were evaluations regarding the emotional state that can be identified threw it, with descriptions like calm, relaxed, cheerful, excited, tense, irritated, sad, and bored. For the evaluation of the interaction between the participants and the humanoid robot, the emotional control system was leveraged [37].

The objective of [19] is to contribute to the development of a social robot that reacts empathetically to users, fostering a positive attitude. Essentially, the proposed system integrates two distinct input devices, namely touch and a camera, to assess users' emotional states and deliver a suitable response. These sensors encompass an emotion recognition system utilizing a trained neural network to identify facial emotions and a fur touch sensor for discerning the specific type of touch applied by the user[19]. Flagg et al., [22] gather information from touch sensors so that machine learning analysis can identify nine fundamental touches that convey emotion. The model's accuracy in determining which participant is touching the prototype was 79% in a study including 16 participants[22]. [21] focus on the design of therapeutic robotic companion and thus, they evaluated the interaction among others based on how pleasant it can be. Additionally, for a changing texture skin robot, it is not only important to evaluate how it is perceived (i.e., liveliness), but also based on the feeling it can trigger, and thus [30] among others evaluated the attachment and other indications of emotional expressiveness.

Robots designed for social interaction often express their inner and emotional states through nonverbal behavior. In most systems, this is accomplished through facial expressions, gestures, movement, and tone of voice. In [29] they propose a new expressive non-verbal channel for social robots in the form of a texture-changing skin. [38]chose to use the iCub robot and leveraged the input data to their method to obtain from the sense of touch, which is an important method for social robots. Emotion on the robot platform is represented by facial expressions, which are supported by a developed control architecture[38]. Tsalamlal et al., [52] present an experiment that highlights how participants combine facial expressions and haptic feedback to perceive emotions when interacting with an expressive humanoid robot. Participants were asked to interact with the humanoid robot through a handshake while looking at its facial expressions. The results showed that they combined facial and tactile cues to assess valence, arousal, and dominance, while they placed more importance on facial expressions when evaluating valence[52]. Yohanan and MacLean [15] examine how people communicate emotional state through touching the Haptic Creature and their expectations of its reactions. A user study is presented where participants selected and performed touches to convey nine different emotions. They describe a "touch dictionary" compiled for their research and present five tentative emotion "intention" categories: protective, comforting, calm, loving, and playful. Chang et al., [18] also worked with the Haptic Creature and investigate the basic principles of emotional touch. This little robot senses the world and communicates its internal state through purring, perking up its ears and regulating its breathing and pulse. These results can help the future design process of social robots by elucidating in detail the needs for an effective human-social robot interaction [15], [18]. [48], after performing three handshakes, asked the participants to rank the handshake based on their preference using three labels “poor”, “moderate” and “good”. The aim of [46] is to investigate whether angry, neutral and happy facial expressions have different effects on tactile responses. Block, et al., [23] examine the realm of emotions conveyed by a social humanoid robot through hugging. They emphasize that hugging is a potent means of experiencing social support and, drawing upon prior research in and beyond HRI, and they put forth six key principles governing the nature of authentic and enjoyable robotic hugs. These principles encompass qualities such as softness, warmth, human-scale dimensions, visual awareness of the user, adaptability in the embrace's size and position, and the ability to discern when the user wishes to conclude the hug, with the use of the robotic platform  HuggieBot 2.0 [23].

In their study, [20] conduct an experimental analysis utilizing touch sensor and accelerometer data, employing a sophisticated machine learning algorithm to discern nine basic emotions. This approach enhances the utility of cost-effective sensors to extract valuable emotional information. Additionally, [10] outline challenges and trends in the field, highlighting the successful utilization of tactile interaction for elevated communication. Their analysis shows a predominant focus on emotions like happiness, sadness, anger, and fear, while noting a relative dearth of emphasis on the communication of disgust and surprise [10].

*3.3.2 Combination of touch and gaze behavior* [8], [36], [42], and [52]

Hirano et al., [8] conduct an experiment involving 28 participants interacting with a Pepper robot, revealing a preference for gaze behavior that focused solely on their faces during touch. In a similar vein, [36] conducted a separate experiment with 20 participants engaging with a robot, exploring various combinations of facial expressions and touch styles. Results indicated that both facial expressions and touch styles played significant roles in influencing perceived emotions. While there seems to be a resemblance between the studies of [8] and [36], [42] delves into the impact of robot gaze on observers, noting its similarity to human gaze effects, positively influencing attitudes and human-robot interaction performance. The robot in [42] initiated direct eye contact, subsequently turning its gaze to activated lights on either side of the table, demonstrating the influence of gaze direction on interaction dynamics. Additionally, in [52] when participants were doing a handshake with the robot were also paying attention on the robot’s facial expressions especially when they were evaluating valence.

*3.3.3 Russell's circumplex model of emotions*: [28], [16] and [29]

Both [28] and [29]utilized Russell's circumplex model of emotion to examine emotions. In this model, the TU control is mapped in two dimensions, with TU-G and TU-S activated on the positive and negative valence sides of an emotion plane, respectively. In the study conducted by [28], participants associated most texture patterns with specific emotions, demonstrating similar distributions across the three modalities. Meanwhile, [16] had participants choose among sixteen emotion labels based on Russell and Ekman's theories for each emotion presented by the Haptic Creature. The findings indicated equivalences between Ekman's and Russell's emotions, except for Ekman's surprise, which did not align with other emotion labels.

*3.3.4 Transfer emotions between users* [25], [33], [34] and [39]

Devices and systems to transfer emotions focus on the feeling and sense of a hug [25], [33], [34], or a kiss [39]. In their examination of the concept of remote kissing, [39]scrutinize the limitations of conventional communication tools in facilitating intimate interactions between individuals separated by distance. To address this gap, the researchers conceived and constructed a haptic apparatus known as Kissenger, derived from the fusion of "kiss" and "messenger." Kissenger is an interactive device meticulously designed to serve as a tangible interface, enabling the transmission of kisses between two individuals who are geographically apart.

### *3.4 Timing*

*3.4.1 Duration and repetition*: [46], [38], [32], [43], [34], [48] ,[52]

In [46], 24 participants used a tactile device that monitored the amount of force applied in Newton units and the duration of the contact in milliseconds to react as quickly as possible to photos of angry, neutral, and cheerful expressions. The findings demonstrated that when people expressed anger, their touch was far more forceful than when they expressed happiness [46]. The duration and the repetition were also one of the foci when having a handshake with a robot [58], and hug applications[34]. Timing and the duration of holding the robot is also a matter of investigation for studies related to hug [32]. [38] include a sequential analysis method that, based on the accumulation of elements from tactile interaction, allows accurate results for touch recognition to be achieved. Finally, in the review of [43] can be found information about the duration and repetition in artificial touch in the human-robot interaction.

*3.4.2 Frequency of texture change*: [28], [29], [30], [47]

In parallel with the model of emotions they apply, both [28] and [29] focus on the frequency of changing texture, while using a surface with small blisters or spikes. The absolute value of the valence dimension is quantified by the amplitude of texture change, while the arousal dimension has been mapped to the frequency of change.

Following the identification of a chemical composition suitable for crafting the robot with variable smoothness, [47]prototyped resembling octopuses were developed. These prototypes featured tentacles with adjustable softness based on touch feedback. User trials were executed to assess participants' ability to detect smoothness changes and to evaluate the viability and potential of this approach.

The characteristics of a material incorporating variable stiffness and tactile components could significantly shape the tactile encounter in interactions. For instance, [30] to intensify an unpleasant sensation, a soft tip on a spike can be substituted with a sharp, rigid element. Regarding its motion, the speed of deformation can be adjusted by modulating the frequency of internal ventilation, while internal air pressure and elastomer elasticity determine the stiffness and range of deformation. Furthermore, a texture unit can be structurally configured to convey diverse forces, offering varied haptic feedback [30].

*3.4.3 Long/short duration of interaction and repeatability*: [50], [27], [9], [44] and [45]

In the duration of interaction and especially on the categorization of it between long and short, [9] observed that female participants expressed emotions for an extended duration, engaging in more diverse interactions and touching a greater number of areas on the robot's body, in contrast to their male counterparts. For the accurate detection and recognition of touch, [50] with the use of microphones and [27], also measure and categorize the duration of touch.

The measurement of long/short duration of interaction together with repeatability was a vital measurement on [44] and [45], repeatedly measuring the touch sensor and accelerometer for building a decision tree based on them.

Table A. Detailed Presentation of Papers used on the PRISMA analysis

| | Title | Technology | Surface Material | Type of Touch | Emotion | Timing (Duration/ Repeatability) |
|---|---|---|---|---|---|---|
| [15]<br><br>[16]<br><br>[18]<br><br>[20] | The role of affective touch in human- robot interaction: Human intent and expectations in touching the haptic creature<br><br>Design and assessment of the haptic creature's affect display<br><br>Gesture recognition in the haptic creature<br><br>Recognizing affect in human touch of a robot | 56 surface mounted force sense resistors (FSR), accelerometer | The Haptic Creature- fur | Manipulations for contact, cradle, finger idly, grap, hold, hug , kiss, lift, massage, nuzzle, pat, pic, poke, press, pull, push, rock, rub, scratch, squeeze, stroke, swing, tap, tickle, toss, tremple | Manipulations for distressed, aroused, excited, miserable, neutral, pleased, depressed, sleepy, relaxed | - |
| [48] | Towards natural handshakes for social robots: human-aware hand grasps using tactile sensors | Sensor hall-effect (magnetic field) | Soft elastomer body (hand) | Measurement of pressure values of sensors in N for men and women | Questionnaire on whether they felt comfortable with the handshake | - |
| [24] | Pressure variation study in human-human and human-robot handshakes: Impact of the mood | Gloves with 23 piezoresistive sensors and accelerometer | Glove made of elastic fabric | Measurement of pressure values of of sensors in kPa for handshake | Manipulations for calm, relaxed, cheerful, excited, tense, irritated, sad, and bored | - |
| [26] | Remote Handshaking: Touch Enhances Video-Mediated Social Telepresence | Not clearly mention the type of touch sensors | Use of urethane gel, urethane sponge and a thin cloth glove | Handshake | - | - |
| [52] | Affective Handshake with a Humanoid Robot: How do Participants Perceive and Combine its Facial and Haptic Expressions? | Meka robot | Meka robot | handshake | Interacting with the robot through a handshake while looking at its facial expressions (valence, arousal, dominance) | detailed presentation of data and manipulations |
| [25] | The Hug: An Exploration of Robotic Form for Intimate Communication | not clearly mention the type of touch sensors | Velvet and silk fabrics. It also has thermal fibers inside | squeeze, stroke, pet | The feelings of one user are transferred through "The Hug" to the other user | - |
| [33] | Hug Over a Distance | not clearly mention the type of touch sensors | A comfortable vest that has an air pump to inflate and a furry koala | Rub (koala) | One user's emotions are transferred from the koala to the other user's vest | - |
| [39] | Mediating intimacy in long-distance relationships using kiss messaging | FSR sensors to detect kiss | A special silicone with RTV 560 was chosen for the lips | Kiss | One user's feelings are transferred through the kiss to the robot to the other user | - |
| [23] | The six hug commandments: design and evaluation of a human-sized hugging robot with visual and haptic perception | Two air chambers form the front and back of the robot's torso | Appropriate clothing is selected to cover the soft internal materials | Behaviors regarding hug has been studied | Manipulations for understood, trust,nice hug, like, good idea, afraid,impressed, cooperate, threatened, useful,helpful, supportive,social | - |
| [32] | Robot reciprocation of hugs increases both interacting times and self-disclosures | Shokkaku Cube (Touchence Inc.) | Human-sized robot teddy bear | Behaviors regarding hug has been studied | - | Interaction time in minutes |
| [17] | Conductive fur sensing for a gesture-aware furry robot | Sense of touch from conductive fur | Fur | Manipulations for stroke,scratch, light touch | - | - |
| [19] | Development of a Multi-sensor Emotional Response System for Social Robots | Touch sensitive fur (fur fabric, conductive yarn and conductive fabric) | Fur | Manipulations for single stroke, multi-stroke, pat, poke, scratch | Manipulations for anger, sadness, neutral, surprise, fear, disgust,combined with camera detecting users’s facial expressions | - |
| [22] | Affective Touch Gesture Recognition for a Furry Zoomorphic Machine | Piezoresistive sensor and conductive fur are used externally | Fur, styrofoam interiors and plastic frame | Stroke, scratch pull, rub, tickle, squeeze, pat, contact, no touch | - | - |
| [8]<br><br><br><br><br>[36] | How do communication cues change impressions of human–robot touch interaction?<br><br>Communication cues in a human-robot | Pepper robot | Pepper robot | Touch to robot and touch by robot | During the interaction, participants prefer the robot to look at them in the face | - |
| | touch interaction | | | | | |
| [31] | A haptic empathetic robot animal for children with autism | Nao robot with attached touch sensors on the hand | Nao dressed in soft koala costume | Pat,hand-rest, squeeze, scratch, poke | - | - |
| [35] | Social touch in human–robot interaction: Robot -initiated touches can induce positive responses without extensive prior bonding | Nao robot | Nao robot | Touch, no touch (touch by robot) | Manipulations for affective trust, perceived trust, help,intimacy, emotional security, and stimulating companionship | - |
| [9] | Affective touch in human–robot interaction: conveying emotion to the Nao robot | Nao robot | Nao robot | - | Study in men and women for fear, anger, disgust, happiness,sadness, gratitude,sympathy, love, means | Average duration for interaction, long/short contact times, control for repetition frequency |
| [41] | An event-triggered low-cost tactile perception system for social robot's whole body interaction | Capacitive technology | ABS<br><br>Materials | Random instant or slow tapping, sliding, poking (finger and palm) | - | - |
| [37] | Maggie: A robotic platform for human-robot social interaction | Capacitive technology | Girl-like robot figure with height 1.35m (no mention to materials) | - | Emotional control system (ECS) | - |
| [44] | A robust online touch pattern recognition for dynamic human-robot interaction | Load transfer touch sensor and accelerometer | Round shape plastic | Recognition accuracy for hit, pat, rub, push | Emotion recognition procedures are mentioned without detailed explanations | Long/short contact duration and repeatability |
| [45] | Online touch behavior recognition of hard-cover robot using temporal decision tree classifier | Load transfer touch sensor and accelerometer | Round shape plastic | Recognition accuracy for hit, pat, rub, push | | Long/short contact duration and repeatability |
| [42] | Embodied social robots trigger gaze following in real-time HRI | Capacitive touch interface with two large buttons for user interface | Copper surface | - | - | - |
| [53] | Interpretation of the modality of touch on an artificial arm covered with an EIT-based sensitive skin | 19 circular electrodes | Soft fabric | Manipulations for tap,pat,push, stroke,scratch, slap,pull,squeeze, no touch | - | Mentions for touch duration without detailed explanations |
| [49] | Touch sensor for social robots and interactive objects affective interaction | Elastosil LR 3162 A/B (piezoresistive properties) | Cine-Skin Silicone | Measurement of sensor pressure values in kPa | - | - |
| [46] | Haptic responses to angry and happy faces | 1.5" Square Force Sense Resistor (FSR) | Sensor not covered with material | Measurement of sensor pressure values in N | Manipulations for happiness, neutral, anger | Average duration in milliseconds |
| [27] | Different strokes and different folks: Economical dynamic surface sensing and affect-related touch recognition | Fabric pressure sensor with EeonTex conductive fabric | Robot that looks like a small animal (fur) | Manipulations for constant, no touch, pat, rub, scratch, stroke, tickle | - | Long/short contact times |
| [50] | Detecting, locating and recognising human touches in social robots with contact microphones | Vibration is transmitted by the robot's shell, where special microphones collect the information | Maggie and Mini Robots (RoboticsLab, Carlos III University, Spain) | Types of touch described: tap, slap, stroke, tickle | - | Reference to Med-long/short contact time |
| [38] | Expressive touch: Control of robot emotional expression by touch | Humanoid robot iCub | Humanoid robot iCub | Manipulations for hard, soft, caress and pinch | Manipulations for happy, shy, angry, disgust | Duration measurement to recognize each type of touch |
| [21] | Design of a therapeutic robotic companion for relational, affective touch | Electric field, temperature and force | Tedy bear (fur) | Manipulations for ratch, slap, pet, pat, rub, squeeze, contact | Manipulations for painful and pleasant | - |
| [28]<br><br>[29] | Using skin texture change to design emotion expression in social robots<br><br>Soft skin texture modulation for social robotics | Pressure sensor | Surface type for skin that changes texture (small blisters or spikes) | - | Russell's cyclic model of emotions | Frequency of texture change |
| [47] | Development of a Variable-Softness Robot by Using Thermoresponsive Hydrogels for Haptic Interaction with Humans | Flexible touch sensors | Heat responsive gel | - | - | Interaction duration in seconds |
| [30] | What Can a Robot’s Skin Be? Designing Texture-Changing Skin for Human-Robot Social Interaction | No touch sensors | Surface type for shape, material, cavity, force, allocation, configuration, | - | Manipulations for expression, concentration, liveliness, attachment, attention | Frequency of texture change |
| | | | analysis and connection | | direction, indicating interaction | |
| [40] | Large-Area Soft e-Skin: The Challenges Beyond Sensor Designs | Challenges in designing soft e-Skin sensors are analyzed | Challenges in soft e-Skin materials are analyzed | - | - | - |
| [43] | Artificial skin and tactile sensing for socially interactive robots: A review | Force sensor technologies (piezoresistive, capacitive, QTC, optical), proximity, dynamic, sensing, temperature, special sensors | Derived from the sensors analysis | - | Mainly refers to the way people feel about each other according to gender and body parts | - |
| [51] | Social touch technology: A survey of haptic technology for social touch | Review of articles related to force, vibration and temperature technologies | Review articles related to contact departments (head, hand, torso, etc.) | Review articles related to types of touch (hug, handshake, kiss etc) | Review of articles related to emotions (greeting, playful, affection etc) | - |
| [10] | Affective haptics: Current research and future directions | - | Review articles related to contact departments (head, hand, torso, etc.) | - | Review of Articles Related to Emotions | - |
| [12] | Step by Step Building and Evaluation of Low-Cost Capacitive Technology Touch System for Human-Social Robot Interaction | Capacitive technology, accelerometer | 3D printed hard shell surface | Manipulation of touch with different parts of the hand (outer palm, 1 finger or 3 fingers, index, middle, ring finger) | - | Repetitive measurements for testing purposes |

## 9. Discussion and Conclusions

The objective of this study was accomplished by the utilization of the dependable Prisma methodology to identify the touch technologies and interactions between humans and social robots. The review encompasses a total of 42 papers. The study design was first examined, and subsequently, the PRISMA flowchart was given, providing a graphical summary of the research process. A tabular format was employed to document the 42 papers encompassed in the evaluation. Table A encompassed details such as the technology employed, touch surface utilized, forms of touch investigated in each study, as well as references to emotions, interaction duration, and repetition. Ultimately, the review encapsulated the salient findings derived from the many studies, highlighting their collective significance.

This study collected and examined the attributes that comprise the tactile interfaces stated in the context of human-robot interaction and found a range of technologies utilized in touch detection, encompassing a variety distinct methods for detecting touch. The accelerometer is a crucial tool for the detection of emotions.

The primary materials identified for user interaction with the entire construct are fur and fabrics. In addition, it is worth noting that in some instances, robots exhibit a resemblance to diminutive plush animals, in contrast to those that possess human-like attributes. The classification of touch is contingent upon a multitude of diverse and influential aspects. The primary determinants encompass the constrained quantity of sensors integrated within the structures, the utilization of designated touch materials, and the intricacy associated with the user's touch behavior. In the context of materials, it is well observed that prioritizing aesthetics and familiarity can have adverse effects on functionality.

Detecting the user's emotions is a complex process because it firstly requires the accurate recording of touch methods by the sensors. Additionally, important factors include the duration, repetition, and frequency of touches. Moreover, there are studies that suggest the use of other means, such as cameras that detect the user's facial expressions. The use of an accelerometer is also an important parameter for identifying the intensity of a motion, whether it is gentle or violent. The duration, repetition, and frequency of touches can provide extremely valuable information about the user's emotions. However, this is a subject of study that is not clearly reported in most studies.

How often and for how long a type of touch occurs is a specific element in developing an accurate emotion detection algorithm.

Through this entire process and the gathered information, it appears that tactile interaction between humans and social robots is a highly complex field that deserves study. Given that social robots are expected to integrate into many aspects of human daily life, it seems that they will make people feel physically and psychologically comfortable around them.

We recognize the fact that more papers regarding this issue have been published after the end date we set, however, although we did not include them on our analysis, we did our best to mention them on the related work.